\documentclass{article}

\usepackage{microtype}
\usepackage{graphicx}
\usepackage{subfigure}
\usepackage{booktabs} % for professional tables
\usepackage{placeins}
\usepackage{multirow} 
\usepackage{mdframed}
\usepackage[most]{tcolorbox}

\usepackage{hyperref}

\usepackage[accepted]{icml2025}

\usepackage{amsmath}
\usepackage{amssymb}
\usepackage{mathtools}
\usepackage{amsthm}

\usepackage[capitalize,noabbrev]{cleveref}

\theoremstyle{plain}

\theoremstyle{definition}

\theoremstyle{remark}

\usepackage[textsize=tiny]{todonotes}

\icmltitlerunning{Caption First, VQA Second: Knowledge Density, Not Task Format, Drives Multimodal Scaling}

\begin{document}

\twocolumn[
\icmltitle{Caption First, VQA Second:\\
Knowledge Density, Not Task Format, Drives Multimodal Scaling}



\icmlsetsymbol{equal}{*}

\begin{icmlauthorlist}
\icmlauthor{Hongjian Zou}{equal,comp}
\icmlauthor{Yue Ge}{equal,sch}
\icmlauthor{Qi Ding}{comp}
\icmlauthor{Yixuan Liao}{comp}
\icmlauthor{Xiaoxin Chen}{comp}
\end{icmlauthorlist}

\icmlaffiliation{comp}{vivo AI Lab, Shenzhen, China}
\icmlaffiliation{sch}{Wuhan University, Wuhan, China}

\icmlcorrespondingauthor{Hongjian Zou}{hongjianzou@gmail.com}

\icmlkeywords{Machine Learning, ICML}

\vskip 0.3in
]



\printAffiliationsAndNotice{\icmlEqualContribution} % otherwise use the standard text.

\begin{abstract}

Multimodal large language models (MLLMs) have achieved rapid progress, yet their scaling behavior remains less clearly characterized and often less predictable than that of text-only LLMs. Increasing model size and task diversity often yields diminishing returns. In this work, we argue that the primary bottleneck in multimodal scaling is not task format, but \textbf{knowledge density} in training data. We first show that task-specific supervision such as Visual Question Answering (VQA) contributes little incremental semantic information beyond image captions: VQA signals can be reconstructed from captions with negligible performance loss. We then demonstrate that increasing knowledge density—through structured caption enrichment and cross-modal knowledge injection—leads to consistent performance improvements across multimodal and downstream benchmarks. Across controlled experiments, performance correlates more strongly with semantic coverage than with task diversity. These findings suggest that current MLLMs fail to scale primarily because training data lacks sufficient knowledge coverage. We advocate for \textbf{knowledge-centric multimodal training} as a principled foundation for scalable multimodal models.

\end{abstract}

\section{Introduction}
\label{sec:introduction}

Multimodal large language models (MLLMs) have achieved rapid progress across vision–language tasks. However, unlike text-only large language models, their scaling behavior remains weaker and less predictable. Modern text LLMs are typically trained on tens of trillions of tokens, whereas current multimodal models rely on substantially smaller multimodal corpora, often on the order of only a few trillion multimodal tokens. Moreover, existing multimodal supervision frequently encodes limited semantic coverage per training sample. As a result, increasing model size, expanding task diversity, or incorporating additional supervision often yields diminishing returns. This raises a fundamental question: \textbf{what drives scaling in multimodal systems?}

A common assumption is that stronger task supervision leads to stronger reasoning ability. In practice, Visual Question Answering (VQA) has become a default form of multimodal supervision for both pretraining and instruction tuning. Originally introduced as a benchmark for visual reasoning, VQA datasets are now widely treated as essential training signals. This implicitly suggests that to improve multimodal reasoning, one must increase task complexity or diversity.

However, closer inspection reveals a different picture. Many VQA questions restate information already present in image captions, and the corresponding answers often require little more than extracting objects, attributes, or simple relations described in natural language. In other words, VQA supervision frequently contributes task formatting rather than new semantic content. This observation suggests that the limitation of current MLLMs may not lie in insufficient task diversity, but in insufficient knowledge coverage.

This insight becomes particularly important when viewed through the lens of scaling. Despite the dominant role of scaling laws in advancing text-only LLMs, similar scaling behavior has not clearly emerged in multimodal models. Modern VLMs are trained on far fewer multimodal tokens than their text-only counterparts, and increasing multimodal data size does not consistently produce proportional performance gains. This discrepancy indicates that multimodal scaling may depend on a different governing factor.

We argue that this factor is \textbf{knowledge density}—the amount and diversity of semantic information contained in training data. Much of today’s multimodal supervision, especially VQA annotations, provides limited incremental knowledge beyond captions. Captions naturally encode objects, relations, contextual cues, and commonsense information, whereas VQA primarily reshapes this knowledge into an interaction format. If knowledge content, rather than task format, determines scaling behavior, then improving multimodal models requires increasing knowledge density rather than increasing task diversity.

This perspective motivates the central thesis of this paper: \textbf{caption supervision is primary, while task-formatted supervision is secondary; the key bottleneck is knowledge density, not task format.}

We validate this thesis in two stages. First, we demonstrate that VQA supervision can be replaced or reconstructed from captions with negligible performance loss, indicating that captions subsume most task-relevant information. Second, we show that increasing knowledge density through cross-modal knowledge injection leads to consistent scaling improvements. Together, these results suggest that current MLLMs fail to scale not because tasks are insufficient, but because training data lacks sufficient knowledge coverage. We advocate for knowledge-centric multimodal training as a principled foundation for scalable multimodal models.

\section{Related Work}
\label{sec:related_work}

\subsection{Multimodal Large Language Models}

Multimodal Large Language Models extend text-only foundation models to jointly process visual and linguistic inputs, enabling unified reasoning across modalities \cite{alayrac2022flamingo, li2023blip, liu2023visual}. Most modern MLLMs adopt a modular architecture that combines a pretrained vision encoder with a large language model through cross-modal alignment modules. Early systems \cite{alayrac2022flamingo, li2023blip} demonstrated that frozen backbones with lightweight connectors can efficiently transfer linguistic priors to multimodal settings, while more recent approaches \cite{chen2024internvl, bai2025qwen2, bai2025qwen3, zhu2025internvl3} move toward large-scale joint pretraining and interleaved image–text modeling.

Representative model families illustrate different design emphases in multimodal learning. The Qwen-VL series \cite{bai2023qwen, wang2024qwen2, bai2025qwen2, bai2025qwen3} integrates strong language backbones with scalable multimodal alignment, emphasizing multilingual capability, instruction following, and deployment efficiency. In contrast, the InternVL family \cite{chen2024far, chen2024internvl, chen2024expanding, zhu2025internvl3, wang2025internvl3} focuses on high-fidelity visual perception and fine-grained cross-modal representation, prioritizing visual tokenization and perception-intensive reasoning. These efforts demonstrate rapid progress in multimodal architecture design and system capability.

Despite these advances, existing MLLMs primarily emphasize architectural scaling and task-level supervision. Comparatively less attention has been devoted to the semantic properties of multimodal training data and their role in scaling behavior. In particular, current pipelines rely heavily on task-formatted supervision such as VQA, while the contribution of semantic coverage and knowledge density remains underexplored.

Our work shifts the focus from architectural scaling to knowledge-centric scaling, examining how semantic density in training data is associated with multimodal model performance.

\subsection{Supervision Paradigms: Captioning vs. Visual Question Answering}

Multimodal training commonly relies on task-structured supervision to align visual and linguistic representations. Among existing paradigms, image captioning and visual question answering (VQA) represent two fundamentally different forms of supervision.

Image captioning provides declarative supervision in which visual content is described through natural language sentences that capture objects, attributes, relations, and scene context. Large-scale caption datasets such as MSCOCO Captions~\cite{lin2014microsoft, chen2015microsoft} enabled early progress in vision–language learning, and caption supervision remains a core component of many modern multimodal training pipelines.

VQA, originally introduced as a benchmark for visual reasoning~\cite{antol2015vqa}, frames supervision as question–answer interactions grounded in images. In recent years, VQA-style supervision has become a dominant training signal in many multimodal systems. Models such as LLaVA~\cite{liu2023visual, liu2024improved}, Qwen-VL~\cite{bai2023qwen}, and InternVL~\cite{chen2024internvl} rely heavily on question–answer style multimodal instruction data, often constructed from VQA datasets or synthetic VQA generation.

The widespread adoption of VQA-style supervision reflects a common assumption in multimodal learning: that stronger task-structured supervision can improve visual reasoning ability. By framing supervision as explicit questions requiring grounded answers, VQA training is often viewed as encouraging models to reason over visual content rather than merely describe it.

However, the distinction between \emph{semantic content} and \emph{interaction format} has received limited attention. In particular, it remains unclear whether task-form supervision introduces additional semantic knowledge beyond declarative descriptions such as captions.

In this work, we revisit this assumption from a data-centric perspective and investigate whether task supervision truly expands the semantic knowledge available to multimodal models.

\subsection{Scaling Laws in Language and Multimodal Models}

Scaling laws have played a foundational role in the development of large language models. Prior work demonstrates that model performance improves predictably as model size, data scale, and compute increase, revealing smooth power-law relationships between resource scaling and task performance~\cite{kaplan2020scaling, hoffmann2022training}. These findings have guided the design of modern foundation models and motivated large-scale pretraining with trillions of tokens.

\begin{table}[hbtp]
\centering
\small
\caption{Training data scale across recent Qwen model generations.}
\label{tab:traing_data_scale_qwen}
\begin{tabular}{lcc}
\toprule
Model Family & Version & PreTraining Tokens \\
\midrule
\multirow{3}{*}{Qwen (text)} 
& Qwen2 & 7T \\
& Qwen2.5 & 18T \\
& Qwen3 & 36T \\
\midrule
\multirow{3}{*}{Qwen-VL (multimodal)} 
& Qwen2-VL & 1.4T \\
& Qwen2.5-VL & 4.1T \\
& Qwen3-VL & 2.2T \\
\bottomrule
\end{tabular}
\end{table}

Text-only LLMs typically exhibit consistent scaling trends in which newer generations are trained on substantially larger token corpora. For example, the Qwen series scales from 7T tokens in Qwen2 to 18T in Qwen2.5 and 36T in Qwen3.

In contrast, multimodal models do not exhibit similarly consistent data-scaling patterns. As shown in Table~\ref{tab:traing_data_scale_qwen}, the Qwen-VL series increases multimodal training data from 1.4T tokens in Qwen2-VL to 4.1T in Qwen2.5-VL, while the newer Qwen3-VL is trained on a smaller corpus of approximately 2.2T tokens. Despite the reduced token scale, newer models may still achieve competitive or improved performance through changes in data composition and training strategy. These observations suggest that raw token count alone does not fully explain performance trends in multimodal systems.

More broadly, scaling behavior in multimodal learning remains less well understood. While model architectures and visual encoders have grown in scale, multimodal datasets are typically much smaller than text corpora, and improvements from increasing data scale are less consistent. Existing work has explored larger image--text corpora and interleaved multimodal documents~\cite{alayrac2022flamingo, peng2023kosmos}, but clear scaling trends comparable to those observed in text-only models have not yet emerged.

Most existing studies focus on resource-centric variables such as parameter count, dataset size, and training compute. Comparatively less attention has been devoted to the semantic properties of multimodal training data. In particular, multimodal supervision is often expanded through additional task annotations (e.g., VQA), implicitly assuming that task diversity drives performance improvements.

In this work, we examine a complementary perspective: whether multimodal scaling is governed by the \emph{semantic coverage} of training data. We investigate whether model performance correlates more strongly with increases in \textbf{knowledge density} than with increases in task diversity.

\section{Task Format vs. Knowledge Content}
\label{sec:task_format}

\subsection{Problem Formulation: Does Task Supervision Expand Knowledge?}

A widely adopted paradigm in multimodal training is to incorporate task-specific supervision—most prominently VQA—to enhance visual reasoning ability. This practice implicitly assumes that supervision contributes additional semantic knowledge beyond image captions and thereby improves scaling behavior.

However, if multimodal scaling is more strongly associated with knowledge density rather than task format, then the critical question is not whether VQA improves performance, but whether it expands the model’s knowledge frontier.
We therefore ask:
Does VQA supervision introduce incremental knowledge beyond captions, or does it primarily reorganize existing information into an interaction format?

Answering this question requires distinguishing between two factors that are often conflated in multimodal training: the \emph{semantic content} contained in the data and the \emph{task format} used to present that information to the model.

\subsection{Controlled Training Experiments}

To evaluate whether task supervision introduces additional knowledge beyond captions, we construct a series of controlled training experiments. All settings use the same image set, identical model architectures, optimization schedules, and total token budgets. The only variable that changes across configurations is the \textbf{type of linguistic supervision} associated with the images.

\paragraph{Model and Training Setup.}
All experiments use a dense multimodal model with approximately 3B parameters. The model consists of a ViT-based vision encoder (0.4B parameters) and a decoder-only Transformer language model (2.5B parameters). The language model has 32 layers with hidden size 2560, feed-forward dimension 6912, and 20 attention heads. The model employs grouped-query attention, RMSNorm normalization, RoPE positional encoding, and SwiGLU activation functions, with a maximum sequence length of 4096.

Training uses Adam with $(\beta_1=0.9,\beta_2=0.98,\epsilon=10^{-6})$, a peak learning rate of $2\times10^{-4}$ followed by rapid fast-decay learning-rate schedule \cite{xiong2025bluelm}, and a global batch size of 1024. Models are trained in bf16 precision for a fixed budget of 4.33B samples (approximately 100B tokens). Unless otherwise specified, architecture, optimization hyperparameters, and training budgets remain identical across all settings.

We use the following notation throughout this section: Caption$_0$ denotes original caption data; VQA$_0$ denotes original VQA data; Caption$_1$ denotes captions generated from VQA images; and VQA$_1$ denotes synthetic VQA pairs reconstructed from Caption$_1$.

\begin{table}[htbp]
\centering
\small
\caption{Training data mixture across experiment settings.}
\label{tab:data_mixture_experiment}
\begin{tabular}{lccc}
\toprule
Data Type & Baseline & Caption-only & Synthetic-VQA \\
\midrule
Caption & 28\% & 45\% & 28\% \\
VQA & 17\% & 0\% & 17\% \\
Pure Text & 40\% & 40\% & 40\% \\
Other & 15\% & 15\% & 15\% \\
\bottomrule
\end{tabular}
\end{table}

\paragraph{Baseline.}
The baseline follows standard multimodal training practice using a mixture of captions, VQA supervision, pure-text data, and other multimodal datasets (e.g., OCR, grounding, and GUI-agent data). The baseline data mixture is shown in Table~\ref{tab:data_mixture_experiment}.

\paragraph{Experiment 1: Caption-Only Variant.}
To remove task-form supervision entirely, all VQA$_0$ pairs are replaced with model-generated captions derived from the same image set. This configuration trains the model using only caption-based supervision together with pure-text and other multimodal data.

\paragraph{Experiment 2: Synthetic-VQA Variant.}
To test whether VQA supervision can be reconstructed from captions, we first generate enriched captions for the original VQA images and then synthesize VQA pairs (VQA$_1$) from these captions using a structured caption-to-question pipeline. The resulting dataset contains Caption$_0$, synthetic VQA$_1$, pure-text data, and other multimodal data.

These configurations allow us to isolate the role of task-form supervision. If models trained with caption-based supervision achieve performance comparable to the baseline, this would suggest that captions already contain most of the semantic information required for VQA-style reasoning.

\subsection{Results}

We report results across four benchmark categories: general multimodal benchmarks, general text benchmarks, business multimodal benchmarks, and business text benchmarks. The business benchmarks serve as practical downstream evaluations and can be viewed as out-of-domain tests relative to the standard academic benchmarks.

\paragraph{General Multimodal Benchmarks.}

\begin{table}[htbp]
\centering
\small
\caption{General multimodal benchmark comparison.}
\label{tab:general_mllm_benchmark}
\resizebox{\linewidth}{!}{
\begin{tabular}{lccc}
\toprule
Benchmark & Baseline & Caption-only & Synthetic-VQA \\
\midrule
MMMU & 0.526 & 0.5099 & \textbf{0.5267} \\
MMBench & \textbf{0.7507} & 0.7000 & 0.7373 \\
MM-Vet & 0.5583 & 0.5650 & \textbf{0.5752} \\
MMStar & \textbf{0.5467} & 0.5444 & 0.5402 \\
MathVista & \textbf{0.6193} & 0.6173 & 0.6113 \\
MathVision & 0.1437 & 0.1582 & \textbf{0.1632} \\
AI2D & 0.6912 & 0.6959 & \textbf{0.6997} \\
MiaBench & 0.8268 & 0.8251 & \textbf{0.8307} \\
OCRBench & 0.7537 & 0.7597 & \textbf{0.7603} \\
\midrule
Average & 0.6018 & 0.5973 (-0.0045) & \textbf{0.6050} (+0.0032) \\
\bottomrule
\end{tabular}
}
\end{table}

%\FloatBarrier

Table~\ref{tab:general_mllm_benchmark}. reports results on representative multimodal reasoning benchmarks, including MMMU, MM-Vet, MMStar, MathVista, MMBench, AI2D, and MiaBench.

Across these benchmarks, the three training configurations produce highly similar results. The baseline achieves an average score of 0.602, while the caption-only configuration obtains 0.597, and the synthetic-VQA variant achieves 0.605. Performance differences across individual benchmarks remain small, typically within one point.

Notably, the caption-only configuration—despite removing all explicit VQA supervision—maintains comparable performance to the baseline across most benchmarks. Meanwhile, the synthetic-VQA configuration, which reconstructs VQA pairs from captions, slightly exceeds the baseline on several tasks such as AI2D and OCRBench.

These results suggest that explicit VQA supervision contributes little additional information beyond what is already encoded in captions.

\paragraph{General Text Benchmarks.}

\begin{table}[htbp]
\centering
\small
\caption{General text benchmark comparison.}
\label{tab:general_llm_benchmark}
\resizebox{\linewidth}{!}{
\begin{tabular}{lccc}
\toprule
Benchmark & Baseline & Caption-only & Synthetic-VQA \\
\midrule
CFBench(PSR) & 0.5400 & \textbf{0.5420} & 0.5413 \\
GPQA-Diamond & 0.6987 & 0.6768 & \textbf{0.6987} \\
MMLU-Pro & \textbf{0.3013} & 0.2987 & 0.2960 \\
\midrule
Average & \textbf{0.5133} & 0.5058 (-0.0075) & 0.5120 (-0.0013) \\
\bottomrule
\end{tabular}
}
\end{table}
%\FloatBarrier

To ensure that replacing VQA supervision does not affect language capability, we further evaluate models on three representative text reasoning benchmarks: MMLU, GPQA, and CFBench.

As shown in Table~\ref{tab:general_llm_benchmark}, across these benchmarks, performance remains largely stable. The baseline achieves an average score of 0.513, compared to 0.506 for the caption-only configuration and 0.512 for the synthetic-VQA configuration. Differences across benchmarks remain small, indicating that caption-based supervision does not significantly impact text reasoning ability.

This result confirms that removing task-formatted supervision in multimodal training does not degrade core language capabilities.

\paragraph{Business Multimodal Benchmarks.}

\begin{table}[htbp]
\centering
\small
\caption{Business multimodal benchmark comparison.}
\label{tab:business_mllm_benchmark}
\resizebox{\linewidth}{!}{
\begin{tabular}{lccc}
\toprule
Benchmark & Baseline & Caption-only & Synthetic-VQA \\
\midrule
Object Recognition & 0.3628 & 0.3617 & \textbf{0.3861} \\
OCRGRD & \textbf{0.7789} & 0.7513 & 0.7258 \\
OCRREC & 0.6927 & \textbf{0.6982} & 0.6672 \\
Doc Understanding (Attr) & 0.2832 & 0.2840 & \textbf{0.3062} \\
Doc Understanding (Table) & \textbf{0.4474} & 0.4341 & 0.4008 \\
Image Description & 0.3628 & \textbf{0.3766} & 0.3698 \\
Image Grounding & 0.1742 & 0.1813 & \textbf{0.1857} \\
Information Extraction & 0.4285 & \textbf{0.4995} & 0.4592 \\
Screen VIVO-Pro & \textbf{0.1232} & 0.1156 & 0.1113 \\
Screen VIVO-v2.1 & \textbf{0.3805} & 0.3333 & 0.2801 \\
\midrule
Average & 0.4034 & \textbf{0.4036} (+0.0002) & 0.3892 (-0.0142) \\
\bottomrule
\end{tabular}
}
\end{table}
%\FloatBarrier

We further evaluate models on a suite of internally curated business multimodal benchmarks, which involve practical tasks such as OCR grounding, document understanding, information extraction, and image grounding.

As shown in Table~\ref{tab:business_mllm_benchmark}, on business multimodal benchmarks, the caption-only configuration maintains performance comparable to the baseline, while the synthetic-VQA variant shows mixed results across tasks.

Because these tasks differ substantially from the academic benchmarks used during model development, they provide a useful out-of-domain evaluation. This observation motivates the knowledge-centric investigation in Section~\ref{sec:knowledge_centric}.

\paragraph{Business Text Benchmarks.}

\begin{table}[htbp]
\centering
\small
\caption{Business text benchmark comparison.}
\label{tab:business_llm_benchmark}
\resizebox{\linewidth}{!}{
\begin{tabular}{lccc}
\toprule
Benchmark & Baseline & Caption-only & Synthetic-VQA \\
\midrule
Summary & 0.7473 & 0.7460 & \textbf{0.7487} \\
Writing & 0.6993 & \textbf{0.7009} & 0.7007 \\
NER & 0.5779 & 0.5821 & \textbf{0.5822} \\
Keywords & 0.6433 & 0.6443 & \textbf{0.6520} \\
\midrule
Average & 0.6670 & 0.6683 (+0.0013) & \textbf{0.6709} (+0.0039) \\
\bottomrule
\end{tabular}
}
\end{table}
%\FloatBarrier

Finally, we evaluate performance on several in-house business text tasks, including summarization, writing, named entity recognition, and keyword extraction.

Results in Table~\ref{tab:business_llm_benchmark} again remain stable across configurations. The baseline achieves an average score of 0.667, compared to 0.668 for the caption-only configuration and 0.671 for the synthetic-VQA configuration. The differences remain small but consistently non-negative.

These results indicate that caption-based supervision does not negatively impact downstream language applications and may even provide slight improvements.

\subsection{Discussion}

Across all benchmark categories, the three supervision strategies produce highly comparable results, with performance differences typically within one point. The caption-only configuration maintains nearly identical performance to the baseline despite removing all VQA supervision, while the synthetic-VQA variant—constructed entirely from captions—achieves similar results to the original VQA-based training.

To better illustrate why this occurs, we present a representative example below.

\begin{figure}[t]
\centering
\includegraphics[width=0.75\linewidth]{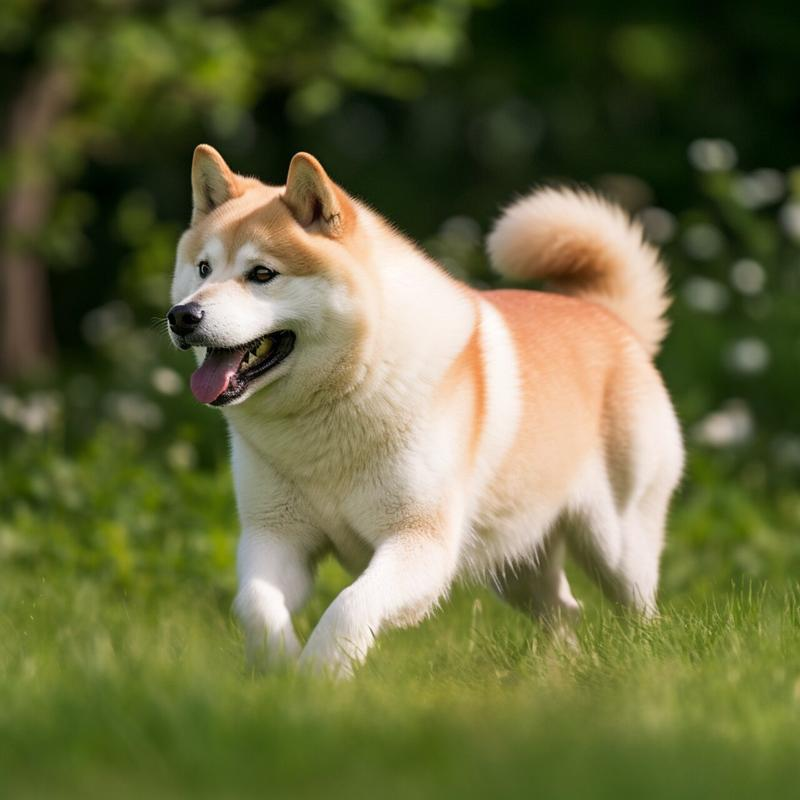}

\vspace{4pt}

\begin{minipage}{0.9\linewidth}
\small
\textbf{Original VQA.}  
Q: What animal is running on the grassland?  
A: Dog.

\textbf{Generated Caption.}  
``A Shiba Inu dog running along the grassland.''

\textbf{Reconstructed VQA.}  
Q: What animal is running along the grassland?  
A: Dog.
\end{minipage}

\caption{Illustrative example showing how the semantic information required for answering the VQA question is already contained in the caption. The VQA pair primarily reorganizes this information into a question–answer format rather than introducing new semantic content.}
\label{fig:vqa_caption_example}
\end{figure}

As illustrated in this example, the generated caption already contains the information required to answer the question. The VQA pair primarily reorganizes this information into a question–answer format rather than introducing new semantic content.

Consistent with this observation, our experiments show that explicit VQA supervision contributes limited additional semantic information beyond what is already encoded in captions. The ability to reconstruct VQA pairs from captions without performance degradation further suggests that captions already contain most of the knowledge required for VQA-style reasoning.

Taken together, these findings indicate that task \textbf{format is not the primary driver of multimodal scaling}. Instead, model capability appears to depend more strongly on the \textbf{semantic coverage of training data} than on the specific supervision format used during training.

This observation motivates the investigation in the next section, where we examine whether explicitly increasing \textbf{knowledge density}—rather than task diversity—can lead to measurable improvements in multimodal model performance.

\section{Knowledge-Centric Multimodal Training}
\label{sec:knowledge_centric}

\subsection{Motivation: Knowledge Density and Multimodal Scaling}

Section~\ref{sec:task_format} showed that modifying task supervision—removing or reconstructing VQA data—does not significantly affect model performance. These results suggest that task format primarily determines the interaction structure between the model and the data, but does not substantially expand the semantic knowledge available during training. If task supervision does not introduce new knowledge, increasing task diversity alone cannot explain improvements in multimodal scaling.

This observation raises a fundamental question: \textbf{what governs scaling behavior in MLLMs?}

We argue that the key factor is the \textbf{knowledge density} of the training data—the amount and diversity of semantic information encoded in each training sample. Unlike text-only LLMs, which benefit from massive token-scale expansion, modern MLLMs are typically trained on comparatively limited multimodal corpora. Caption and VQA datasets often describe visible objects and simple relations, but contain limited contextual, comparative, or long-tail knowledge. As a result, increasing supervision through additional task annotations may reorganize existing information without meaningfully expanding semantic coverage.

Intuitively, scaling improves model capability only when the training data expands the model’s \emph{knowledge frontier}—introducing new objects, relations, contextual facts, and conceptual distinctions. If additional supervision merely reformats existing information into new task structures, scaling gains should remain limited.

Based on this observation, we hypothesize that \textbf{multimodal scaling is governed primarily by knowledge density rather than task diversity.}

To investigate this hypothesis, we develop a knowledge-centric data construction strategy that increases semantic coverage while keeping the task format unchanged. In particular, we introduce a cross-modal knowledge injection pipeline that enriches multimodal training samples with additional relational and conceptual information.

\begin{figure}
    \centering
    \includegraphics[width=1\linewidth]{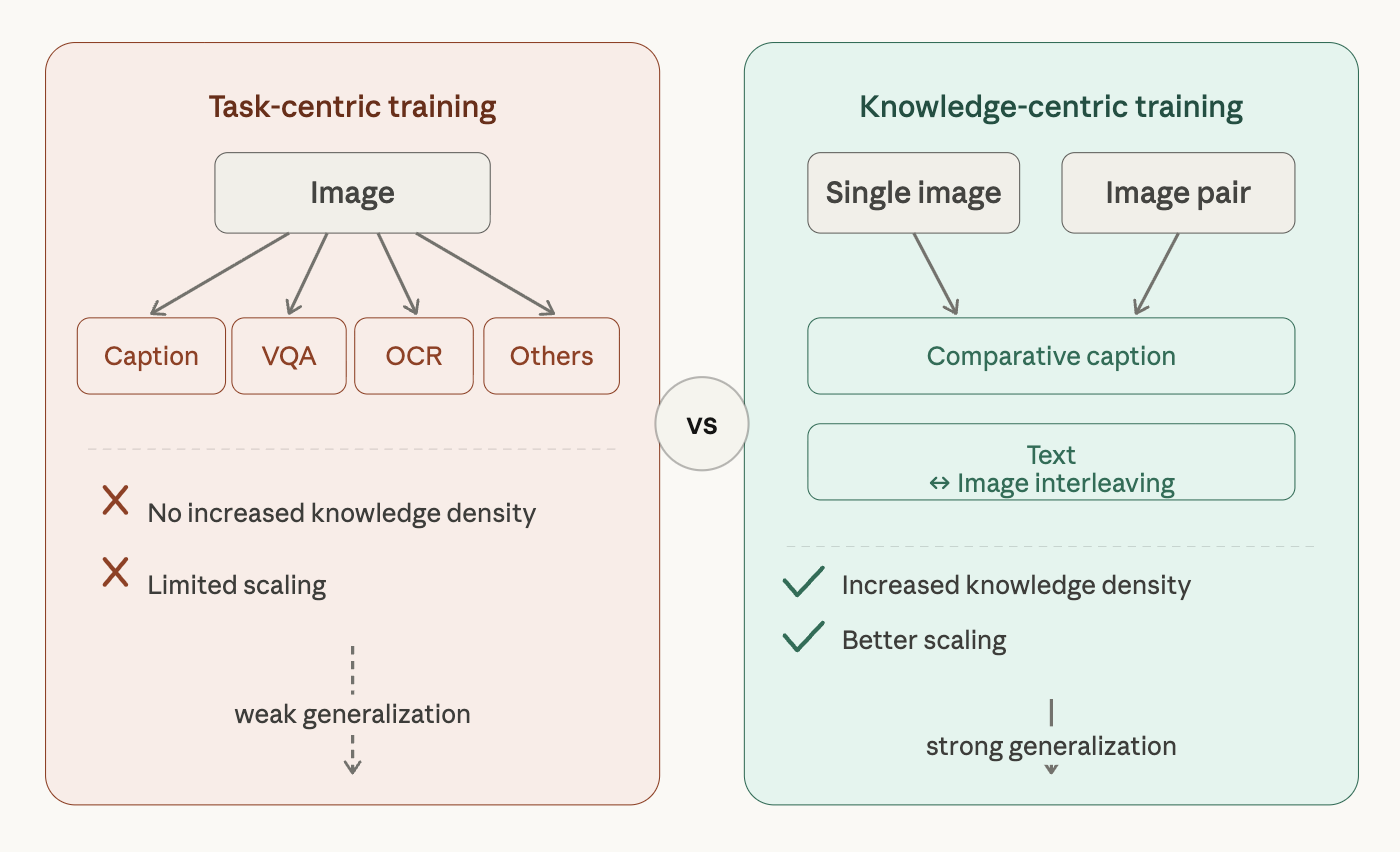}
    \caption{
Conceptual comparison between task-centric multimodal supervision and knowledge-centric training. Traditional pipelines increase task diversity through multiple supervision formats (e.g., captioning, VQA), while our approach increases semantic coverage by constructing knowledge-rich multimodal data. }
    \label{fig:task_centric_vs_knowledge_centric}
\end{figure}

\subsection{Knowledge-Density Interventions via Cross-Modal Knowledge Injection}

\subsubsection{Image Pair Construction}

\begin{figure}
    \centering
    \includegraphics[width=1\linewidth]{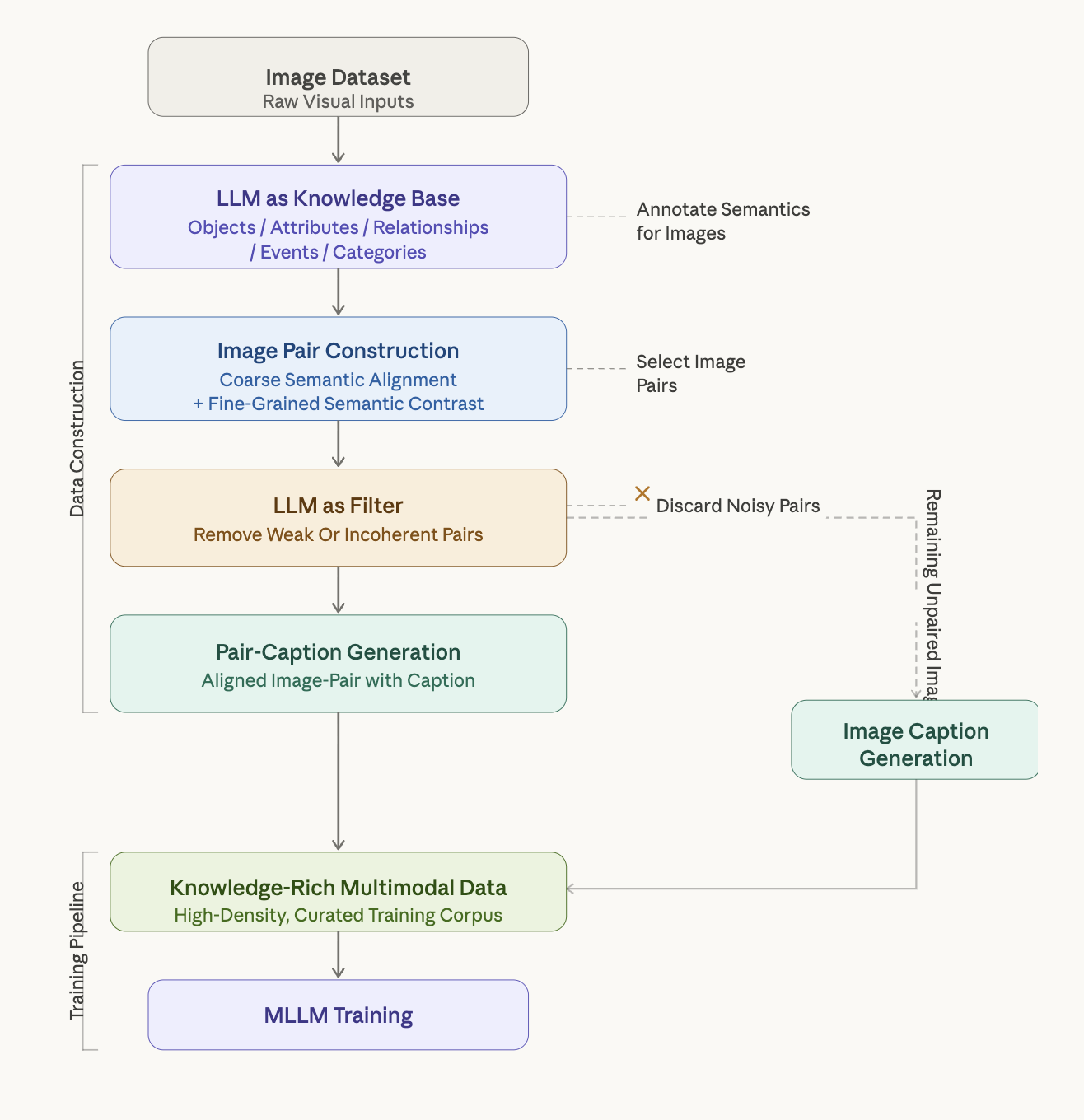}
    \caption{Data construction pipeline for knowledge-centric multimodal training. An MLLM annotates semantic attributes for each image; images are then paired by semantic similarity and contrast, filtered for coherence, and captioned. Unpaired images follow a separate caption generation branch. Both streams yield a knowledge-rich corpus for MLLM training.}
    \label{fig:pair_caption_pipeline}
\end{figure}

To increase knowledge density in multimodal training data, we introduce a structured data construction pipeline that injects additional semantic relationships between images. The key idea is to augment conventional image–caption supervision with comparative and relational knowledge derived from semantically related image pairs.

Unlike standard caption datasets where each image is treated independently, our approach constructs pairs of images that share high-level semantic categories while differing in fine-grained attributes or contextual properties. Training on such paired samples encourages the model to learn richer conceptual distinctions and relational structures, thereby expanding the semantic coverage of the training corpus.

A central challenge is ensuring that the constructed image pairs introduce meaningful knowledge rather than arbitrary visual co-occurrences. Naively pairing images may produce weak or unintuitive relationships that provide little useful supervision.

To address this issue, we design a two-stage pipeline that uses LLMs as both a semantic knowledge base and a semantic validity filter. The LLM first extracts structured semantic descriptors for each image to guide image pairing, and then evaluates the resulting pairs to ensure that the relationships between images are coherent and informative.

\paragraph{LLM as Knowledge Base}

We first use a multimodal large language model to extract structured semantic information from each image–caption sample. For every image, the model generates semantic descriptors including:

\begin{itemize}

\item coarse semantic category (e.g., animal, vehicle, food)

\item fine-grained attributes and subcategories

\item key entities and salient objects

\end{itemize}

These descriptors provide a lightweight knowledge representation summarizing the semantic content of each image.

\paragraph{Image Pair Construction}

Using this representation, candidate image pairs are constructed according to two complementary principles:

\begin{itemize}

\item \textbf{Coarse semantic alignment}: paired images share similar high-level categories.

\item \textbf{Fine-grained semantic contrast}: paired images differ in attributes, subcategories, or contextual properties.

\end{itemize}

This pairing strategy exposes the model to both semantic commonality and discriminative distinctions within the same conceptual space, thereby increasing relational and conceptual coverage in the training data.

We refer to this stage as \textbf{LLM as Knowledge Base}, where the language model serves as a structured semantic organizer for multimodal data construction.

\paragraph{LLM as Semantic Filter}

Naively pairing images may produce weak, trivial, or accidental relationships. To ensure that the constructed pairs provide meaningful supervision, we introduce a semantic filtering stage using a large multimodal model.

For each candidate image pair, the model evaluates whether the relationship between the two images forms a coherent and informative comparison. In particular, the model is prompted to determine whether the relationship could be clearly explained to a young child in simple terms. Pairs that fail this criterion are discarded.

This filtering step removes noisy or unintuitive pairings and preserves semantically informative relationships, ensuring that the constructed training samples contain meaningful conceptual connections rather than incidental visual similarities.

We refer to this stage as \textbf{LLM as Filter}, where the language model acts as a semantic quality controller during data construction.

\subsubsection{Multi-Image Interleaved Descriptions}

To further increase knowledge density beyond single-image descriptions, we construct interleaved data by aggregating multiple semantically related images into a unified textual description.

Specifically, instead of inserting images into pre-existing text documents, we sample groups of three or more related images and prompt a large language model to generate a coherent, long-form description that integrates information across all images. This process encourages the model to synthesize shared themes, relationships, and complementary details, resulting in a denser and more structured semantic representation.

Compared to single-image captions, such multi-image descriptions introduce cross-instance relationships and higher-level abstractions, thereby increasing the overall knowledge density of each training sample.

\subsection{Training Experiment}
\label{subsec:kd_experiment}

We evaluate the proposed knowledge-density interventions using a controlled training setup. All experiments use the same model architecture, optimization configuration, and training budget to ensure fair comparison.

The base model is a multimodal transformer trained with approximately 50B training tokens. Training follows a standard autoregressive objective over mixed multimodal and text data. Unless otherwise specified, all experimental configurations follow those described in Section~\ref{sec:task_format}.

We compare four training configurations:

\paragraph{Baseline.}
The original training corpus containing caption data, VQA supervision, other multimodal data, and pure-text data.

\paragraph{Experiment 1: Pair-Caption-v1.}
Semantically paired image–caption samples replace the original caption and VQA supervision.

\paragraph{Experiment 2: Interleaved.}
Interleaved image–text documents replace the original caption and VQA supervision.

\paragraph{Experiment 3: Pair-Caption-v2.}
Semantically paired image–caption samples replace the original caption data while keeping the original VQA supervision unchanged.

All configurations use identical training schedules and token budgets. Performance is evaluated on both academic benchmarks and proprietary real-world business benchmarks covering multimodal reasoning, text reasoning, and downstream application tasks.

\subsection{Results}
\label{subsec:kd_experiment_result}

We evaluate the proposed knowledge-density interventions across four benchmark categories: general multimodal benchmarks, general text benchmarks, business multimodal benchmarks, and business text benchmarks.

\paragraph{General Multimodal Benchmarks.}

\begin{table}[htbp]
\centering
\small
\caption{General multimodal benchmark comparison.}
\label{tab:kd_general_mllm_benchmarks}
\resizebox{\linewidth}{!}{
\begin{tabular}{lcccc}
\toprule
Benchmark & Baseline & Pair-Caption-v1 & Interleaved & Pair-Caption-v2 \\
\midrule
MiaBench & 0.8317 & 0.8458 & 0.8409 & \textbf{0.8507} \\
MMMU & 0.5124 & 0.5171 & \textbf{0.5210} & 0.5086 \\
MMBench & \textbf{0.7240} & 0.6973 & 0.6693 & 0.7227 \\
MM-Vet & 0.5497 & 0.5526 & 0.5298 & \textbf{0.6076} \\
MMStar & 0.5293 & 0.5256 & 0.4927 & \textbf{0.5375} \\
MathVision & 0.1494 & \textbf{0.1608} & 0.1565 & 0.1595 \\
MathVista & \textbf{0.6007} & 0.5747 & 0.5530 & 0.5870 \\
AI2D & 0.6993 & 0.6942 & 0.6694 & \textbf{0.7047} \\
OCRBench & 0.7380 & \textbf{0.7550} & 0.7163 & 0.7433 \\
\midrule
Average & \textit{0.5927} & 0.5915 & 0.5721 & \textbf{0.6024} \\
\bottomrule
\end{tabular}
}
\end{table}
%\FloatBarrier

Table~\ref{tab:kd_general_mllm_benchmarks} reports performance on representative multimodal benchmarks, including MMMU, MM-Vet, MMStar, MathVista, MMBench, AI2D, and MiaBench.

Among the evaluated configurations, \textbf{Pair-Caption-v2 achieves the strongest overall performance}, improving the average score from 0.593 (baseline) to 0.602. Improvements are observed on several benchmarks, including MM-Vet, MMStar, and AI2D, while performance remains comparable to the baseline on the remaining tasks.

In contrast, the Interleaved configuration consistently performs worse than the baseline across most benchmarks. This suggests that the current interleaved data construction—based on aggregating multiple images into long-form descriptions—may introduce less precise visual–text alignment and diluted semantic signals, limiting its effectiveness in expanding semantic coverage.

Overall, these results indicate that \textbf{structured semantic pairing of images can improve multimodal reasoning performance}, whereas unstructured multimodal augmentation may not yield consistent benefits.

\paragraph{General Text Benchmarks.}

\begin{table}[htbp]
\centering
\small
\caption{General text benchmark comparison.}
\label{tab:kd_general_llm_benchmarks}
\resizebox{\linewidth}{!}{
\begin{tabular}{lcccc}
\toprule
Benchmark & Baseline & Pair-Caption-v1 & Interleaved & Pair-Caption-v2 \\
\midrule
MMLU-Pro & 0.3097 & 0.3233 & \textbf{0.3330} & 0.3180 \\
GPQA-Diamond & 0.6414 & 0.6532 & 0.6802 & \textbf{0.7104} \\
CFBench(PSR) & 0.5363 & 0.5507 & 0.5490 & \textbf{0.5553} \\
\midrule
Average & 0.4958 & 0.5091 & \textit{0.5207} & \textbf{0.5279} \\
\bottomrule
\end{tabular}
}
\end{table}
%\FloatBarrier

We further evaluate text-only reasoning ability on three representative benchmarks: MMLU-Pro, GPQA-Diamond, and CFBench.

As shown in Table~\ref{tab:kd_general_llm_benchmarks}, all knowledge-density interventions improve text performance relative to the baseline. The Pair-Caption-v2 configuration achieves the largest improvement, increasing the average score from 0.496 to 0.528.

Notably, the GPQA benchmark exhibits a particularly large gain, suggesting that knowledge-rich multimodal supervision may improve the model's ability to perform complex reasoning. These results indicate that increasing multimodal knowledge density can positively transfer to text reasoning capability.

\paragraph{Business Multimodal Benchmarks.}

\begin{table}[htbp]
\centering
\small
\caption{Business multimodal benchmark comparison.}
\label{tab:kd_business_mllm_benchmarks}
\resizebox{\linewidth}{!}{
\begin{tabular}{lcccc}
\toprule
Benchmark & Baseline & Pair-Caption-v1 & Interleaved & Pair-Caption-v2 \\
\midrule
OCRGRD & 0.5957 & 0.4656 & 0.4982 & \textbf{0.6651} \\
OCRREC & \textbf{0.6850} & 0.6697 & 0.6341 & 0.6676 \\
Screen-VIVO-Pro & 0.0875 & \textbf{0.1983} & 0.0908 & 0.1918 \\
Screen-VIVO-v2.1 & 0.2813 & \textbf{0.4916} & 0.3171 & 0.4626 \\
Object Recognition & \textbf{0.3011} & 0.2483 & 0.2111 & 0.2872 \\
Info Extraction & 0.7467 & 0.7738 & 0.7463 & \textbf{0.7749} \\
Doc Understanding(Attr) & \textbf{0.3038} & 0.2889 & 0.2864 & 0.2901 \\
Doc Understanding(Table) & 0.4180 & 0.3761 & 0.4283 & \textbf{0.4318} \\
Image Description & 0.3770 & \textbf{0.3894} & 0.3775 & 0.3867 \\
Image Grounding & 0.1713 & \textbf{0.2533} & 0.1927 & 0.2409 \\
\midrule
Average & 0.3967 & \textit{0.4155} & 0.3782 & \textbf{0.4399} \\
\bottomrule
\end{tabular}
}
\end{table}
%\FloatBarrier

To evaluate practical multimodal capability, we test the models on a suite of business-oriented benchmarks involving OCR grounding, document understanding, information extraction, and screen understanding.

As shown in Table~\ref{tab:kd_business_mllm_benchmarks}, the Pair-Caption-v2 configuration achieves the best overall performance, improving the average score from 0.397 (baseline) to 0.440. Gains are observed on several tasks, including OCR grounding, screen understanding, and information extraction.

Because these tasks differ substantially from the academic benchmarks used during model development, they provide a useful out-of-domain evaluation. The improvements suggest that increasing knowledge density can enhance real-world multimodal generalization.

\paragraph{Business Text Benchmarks.}

\begin{table}[htbp]
\centering
\small
\caption{Business text benchmark comparison.}
\label{tab:kd_business_llm_benchmarks}
\resizebox{\linewidth}{!}{
\begin{tabular}{lcccc}
\toprule
Benchmark & Baseline & Pair-Caption-v1 & Interleaved & Pair-Caption-v2 \\
\midrule
Writing Bench & 0.6165 & 0.6369 & 0.6262 & \textbf{0.6369} \\
Summary & 0.7507 & \textbf{0.7597} & 0.7507 & 0.7510 \\
NER & 0.5994 & 0.6077 & \textbf{0.6152} & 0.5920 \\
Keywords & 0.6559 & \textbf{0.6847} & 0.6531 & 0.6755 \\
\midrule
Average & 0.6556 & \textbf{0.6722} & 0.6613 & \textit{0.6638} \\
\bottomrule
\end{tabular}
}
\end{table}
%\FloatBarrier

Finally, we evaluate performance on several business text tasks, including summarization, writing, named entity recognition, and keyword extraction.

As shown in Table~\ref{tab:kd_business_llm_benchmarks}, all training variants outperform the baseline. The Pair-Caption-v1 configuration achieves the highest overall score (0.672), while the other variants also produce modest improvements over the baseline.

These results indicate that knowledge-centric multimodal training does not degrade downstream language capability and may provide small but consistent improvements for practical text applications.

\paragraph{Summary}

Across benchmark categories, \textbf{Pair-Caption-v2 provides the strongest multimodal performance}, while Pair-Caption-v1 performs best on several business text tasks. 

Overall, these results suggest that \textbf{structured knowledge augmentation through semantically paired multimodal data can improve both multimodal reasoning and downstream task performance}, providing a more effective alternative to simply increasing task-formatted supervision.

\subsection{Knowledge Density Analysis}

To better understand the source of the observed improvements, we analyze the semantic information contained in the training data.

We define the \textbf{knowledge density} of a training sample as the number of distinct semantic elements it contains. These elements include objects, attributes, relations, events, and contextual facts extracted from the image--text pair.

\[
\mathrm{KD}(x) = |K(x)|
\]

where $K(x)$ denotes the set of extracted semantic elements in sample $x$.

To estimate this quantity, we use a large language model as a semantic analyzer. For each training instance, the model is prompted to extract structured semantic elements describing the visual scene and its relationships. The total number of extracted elements is used as a proxy for the knowledge content of that sample.

We note that our knowledge-density estimate is an approximate proxy based on LLM-assisted semantic extraction rather than a complete formal measure of information content.

Importantly, knowledge density does not simply correspond to caption length. Many paired captions contain similar token counts as single-image captions but introduce additional relational or comparative semantic elements, which increases the number of distinct knowledge elements extracted by the analyzer.

We compare the original caption/VQA supervision with captions constructed from semantically paired images (Pair-Caption-v2). The analysis reveals a substantial increase in semantic information density. Figure~\ref{fig:kd_compare} shows that paired-image captions contain on average 32 knowledge elements per sample, compared with 22 for conventional caption/VQA supervision, corresponding to a 45\% increase.

This increase arises because paired-image descriptions naturally introduce comparative and relational information between images. Compared with single-image captions, paired descriptions frequently include contrasts, attribute comparisons, and contextual explanations, thereby enriching the semantic structure encoded in each training sample.

Combined with the performance improvements observed in Section~\ref{subsec:kd_experiment} and Section~\ref{subsec:kd_experiment_result}, these findings provide empirical evidence supporting our central hypothesis: increasing the semantic knowledge density of multimodal training data can lead to improved model capability.

\iffalse

\begin{table}[h]
\centering
\small
\caption{Estimated semantic knowledge elements per training sample.}
\begin{tabular}{l c}
\toprule
Data Source & Knowledge Elements per Sample \\
\midrule
Original Caption / VQA & 22 \\
Paired Image Caption & 32 \\
\bottomrule
\end{tabular}
\end{table}

\fi

\iffalse
\begin{figure}
    \centering
    \includegraphics[width=1\linewidth]{kd_compare.png}
    \caption{Comparison of estimated knowledge density per training sample.}
    \label{fig:kd_compare}
\end{figure}
\fi

\begin{figure}
    \centering
    \includegraphics[width=1\linewidth]{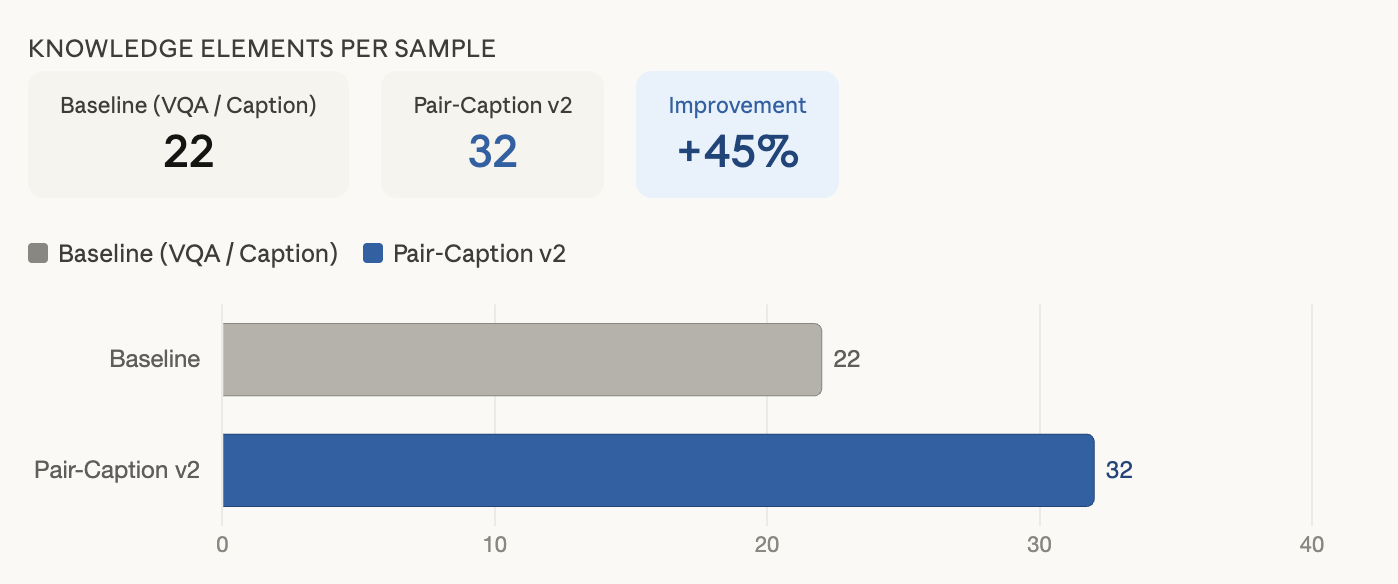}
    \caption{Comparison of knowledge density between training paradigms. Pair-Caption-v2 samples yield an average of 32 knowledge elements per sample, compared to 22 for standard VQA\/Caption supervision — a 45\% increase. Knowledge elements are extracted by prompting an LLM to identify objects, attributes, relationships, events, and categories present in each training sample.}
    \label{fig:kd_compare}
\end{figure}

\subsection{Discussion and Implications}

\paragraph{Knowledge as a Governing Variable.}
Across controlled experiments, we observe a consistent pattern: performance improvements emerge when the semantic coverage of the training data increases, and remain largely unchanged when supervision formats vary without expanding knowledge content.

This observation suggests that \textbf{knowledge density plays a central role in governing multimodal scaling behavior}. Increasing task diversity alone—without introducing new semantic information—does not produce systematic improvements. In contrast, enriching captions and introducing structured cross-modal relationships lead to consistent performance gains.

Importantly, this does not imply that task supervision is unimportant. Instead, our results indicate that \textbf{task format primarily shapes the interaction structure between the model and the data, while the semantic coverage of the data determines the representational capacity that the model can acquire}. When knowledge density remains unchanged, altering the task format yields only limited improvements.

From this perspective, the key scaling question shifts from \textit{how many tasks a model is trained on} to \textit{how much semantic coverage the training data provides}.

\paragraph{Implications for Multimodal Scaling.}

Text-only large language models exhibit predictable scaling behavior as token count and knowledge accumulation increase. Our results suggest that multimodal scaling follows a related but distinct principle: performance improvements correlate more strongly with increases in \textbf{semantic coverage} than with increases in \textbf{task diversity}.

In practice, this means that scaling multimodal models may depend primarily on expanding the model’s \textbf{knowledge frontier}—its coverage of objects, relations, contextual information, and cross-modal grounding—rather than simply increasing the number of supervision tasks.

This perspective provides a potential explanation for the comparatively weak scaling behavior observed in many current multimodal models. Although task diversity has expanded rapidly in recent years, the underlying multimodal corpora remain relatively modest in scale and limited in semantic richness compared to large text corpora.

We note that knowledge density is likely a \textbf{necessary but not sufficient} condition for scaling. Higher-level abilities such as reasoning, abstraction, and compositional generalization may require additional architectural or training innovations. However, without sufficient semantic coverage, such capabilities are unlikely to emerge reliably.

Overall, these findings suggest that future progress in multimodal scaling may depend less on designing new supervision tasks and more on constructing \textbf{knowledge-rich multimodal corpora} that expand semantic coverage across visual and textual domains.

\section{Conclusion}
\label{sec:conclusion}

In this work, we investigated the role of task supervision and semantic coverage in training multimodal large language models. Our results challenge the widely held assumption that increasing task diversity—particularly through VQA-style supervision—is the primary path toward stronger multimodal reasoning.

In Section~\ref{sec:task_format}, we showed that task supervision contributes limited incremental knowledge beyond captions. Removing VQA supervision does not degrade performance, and VQA pairs reconstructed from captions achieve comparable results. These findings indicate that captions already encode most of the semantic information required for VQA-style reasoning.

In Section~\ref{sec:knowledge_centric}, we introduced the concept of \textbf{knowledge density} as a proxy for the semantic coverage of training data. Through structured cross-modal knowledge injection and paired-caption data construction, we demonstrated that increasing knowledge density leads to consistent performance improvements while keeping the task format unchanged.

Taken together, these results suggest that \textbf{multimodal scaling is governed more strongly by knowledge density than by task diversity}. Rather than relying on increasingly complex task annotations, future multimodal training pipelines may benefit from focusing on knowledge-rich supervision that expands semantic coverage across visual and textual domains.

More broadly, our findings motivate a shift toward \textbf{knowledge-first multimodal supervision}, where semantically dense image–text corpora serve as the foundation of pretraining, and task-specific formats are introduced primarily during alignment or downstream adaptation.

\section{Future Work}
\label{sec:future_work}

While our results highlight the importance of knowledge density in multimodal scaling, several directions remain for future research.

First, more systematic methods for knowledge-centric data construction should be explored. In this work, we introduced semantic image pairing and interleaved image–text data construction as practical mechanisms for increasing knowledge density. Future work may investigate automated knowledge extraction, large-scale semantic retrieval, and structured knowledge augmentation for multimodal corpora.

Second, large-scale interleaved image–text corpora may provide a more natural foundation for multimodal scaling. Constructing multimodal datasets with richer relational structure, broader long-tail coverage, and stronger cross-modal grounding may enable more predictable scaling behavior.

Third, while knowledge density appears to be an important factor in multimodal scaling, it is unlikely to be the only one. Future work should investigate how semantic coverage interacts with reasoning, abstraction, and compositional generalization in multimodal models.

Finally, developing a more formal theory of multimodal scaling remains an open challenge. Understanding how knowledge density, data scale, and model capacity jointly influence generalization could provide a more principled foundation for designing next-generation multimodal foundation models.


\bibliography{example_paper}
\bibliographystyle{icml2025}

%%%%%%%%%%%%%%%%%%%%%%%%%%%%%%%%%%%%%%%%%%%%%%%%%%%%%%%%%%%%%%%%%%%%%%%%%%%%%%%
%%%%%%%%%%%%%%%%%%%%%%%%%%%%%%%%%%%%%%%%%%%%%%%%%%%%%%%%%%%%%%%%%%%%%%%%%%%%%%%
% APPENDIX
%%%%%%%%%%%%%%%%%%%%%%%%%%%%%%%%%%%%%%%%%%%%%%%%%%%%%%%%%%%%%%%%%%%%%%%%%%%%%%%
%%%%%%%%%%%%%%%%%%%%%%%%%%%%%%%%%%%%%%%%%%%%%%%%%%%%%%%%%%%%%%%%%%%%%%%%%%%%%%%
\newpage
\appendix
\onecolumn

\section{Appendix}

\subsection{Prompts for Training Data Construction}

\subsubsection{Prompt for Single-Image Caption Generation}

The following prompt is used to generate image captions in our data construction pipeline.

\begin{mdframed}
\small
\ttfamily

Please generate an objective and complete English caption for the input image, focusing on the main content.

[Scope and Priority of Information]

Focus the description on the main content area that conveys the core information in the image, such as text, main titles, key data, or critical information blocks.

For web pages or interface screenshots, do not describe or mention the following areas unless they themselves constitute the main information:

\begin{itemize}
    \item top navigation bar or menu bar
    \item footer area
    \item record-filing information, copyright notices, or legal terms
    \item auxiliary functional areas, statistical information, or decorative elements
    \item entry statistics at the bottom of the page, such as view count or edit count
    \item bottom links such as ``Getting Started,'' ``I have questions,'' or ``Feedback and Complaints''
    \item links such as ``Deregistration Information'' or ``National Enterprise Credit Information Publicity System''
\end{itemize}

[Mandatory Requirements]

Within the main content area, describe all identifiable information as completely as possible, including text, objects, structures, and their relationships.  

For text in the main content area, transcribe it verbatim. Do not use generalized or omitted expressions such as ``for example,'' ``including,'' or ``etc.''  

Clearly describe spatial relationships among core elements, such as ``on the left,'' ``at the top right,'' ``in front of,'' ``behind,'' ``adjacent to,'' ``surrounding,'' or ``located above.''  

Clearly describe hierarchical relationships or correspondences among elements.  

If there are interactions or actions, describe who is doing what to what.  

If the image is a map, describe directions, orientations, and positional relationships among landmarks.

[Prohibited]

Do not infer unseen information such as identity, emotions, or intentions.  

Do not evaluate the value or quality of the content.  

Do not describe areas explicitly listed above as ignored by default.  

Output a continuous and natural English paragraph describing only the objective facts in the main content area of the image. Do not mention auxiliary areas and do not provide a summary at the end.

Please output the English caption:

\end{mdframed}

\subsubsection{Prompt for Caption-to-VQA Generation}

The following prompt is used to generate VQA pairs from image captions in our data construction pipeline.

\begin{mdframed}
\small
\ttfamily
Please generate high-quality English Visual Question Answering (VQA) data based on the given [Image Description]. All questions and answers must be strictly grounded in the explicit, objective information provided in the image description. Do not introduce any subjective judgments, speculative content, or information not present in the description.

[General Constraints (Applicable to All Q\&A)]

1. All questions must be answerable with a single, clear, and unambiguous answer that can be directly derived from the image description, without relying on common-sense supplementation or implicit reasoning.

2. The only allowed information types are:

-- explicitly mentioned objects  

-- directly observable attributes (such as color, shape, material, state, etc.)  

-- explicitly stated quantities, positional relationships, spatial relationships, or interaction relationships (e.g., ``next to \ldots'', ``in front of \ldots'')

3. The following content is strictly prohibited:

-- emotions, atmosphere, or aesthetic evaluations  

-- intentions, psychological states, or future behaviors  

-- uncertain expressions (such as ``may,'' ``seems,'' ``speculates,'' etc.)

4. For web pages or interface screenshots, do not answer questions about the following information by default, unless they themselves constitute the main content:

-- top navigation bar or menu bar  

-- footer area  

-- registration information, copyright notices, or legal disclaimers  

-- auxiliary function areas, statistics, or decorative elements  

-- the ``entry statistics'' area at the bottom of the page, such as ``view count'' and ``edit count''  

-- links at the bottom of the page, such as ``Getting Started,'' ``I have a question,'' ``Complaints and Suggestions''  

-- link contents such as ``Registration Information'' or ``National Enterprise Credit Information Publicity System''  

-- reference links or reference dates  

-- entry galleries or entry metadata  

-- buttons such as ``Broadcast,'' ``Discuss,'' or ``Upload Video''  

-- interactive icons such as ``Favorite,'' ``Share,'' or ``Like''  

-- editing prompt content such as ``This entry lacks an information section. Add relevant content to make the entry more complete and quickly level up. Start editing now!''

[Number and Structure Requirements]

1. The total number of Q\&A pairs should be controlled between 5 and 15, prioritizing quality over quantity.  

2. Questions must include both global questions and detail-oriented questions, with detail-oriented questions significantly outnumbering global questions.

[Global Question Requirements]

1. Include at least one global question that prompts an overall description of the main content of the entire image.  

2. The questions should be general and summarizing, with answers that can be understood as a concise version of the image description.  

3. Answers should be one natural paragraph, shorter than the original description but significantly longer than a single-sentence summary, avoiding bullet-point or itemized format.

[Detail Question Requirements]

1. Detail questions should target specific and explicit details in the image description, such as:

-- attributes, states, or quantities of individual objects  

-- clearly described spatial positions or relative relationships  

-- explicitly appearing interactions between people or objects  

2. Detail questions should aim to cover different key information points in the image description, without requiring exhaustive coverage of all details.  

3. Detail questions should prioritize information that is valuable for understanding the image content and avoid unnecessary or low training-value details. For page-like images, focus on the main text or core information rather than layout or decorative elements.  

4. Different detail questions should correspond to different information points or attribute dimensions, avoiding repetition or high overlap.  

5. Answers to detail questions must be accurate and concise, mainly consisting of nouns, quantities, or explicit relationships, without additional explanation, modification, or inference.  

6. Avoid generating the following types of detail questions:

-- repetitive or highly similar questions  

-- low-value questions generated merely to increase quantity  

[Output Format]

Please output a JSON list in which each element contains ``question'' and ``answer'' fields:

[{"question": "specific question", "answer": "corresponding answer"}]  
Please return only the JSON list, without any additional explanation, title, or commentary.

[Image Description]

\{cap\}

[Example]

Image description:  

``A cactus approximately 60 centimeters tall, with a green cylindrical stem covered in prominent spines. At the top, three small pink flowers are blooming, with petals spreading outward and yellow stamens. The cactus grows in a vast desert, where the sand is light yellow and scattered with rocks of varying sizes. A rock near the cactus is gray, with a rough surface and visible cracks. In the distance, gently rolling sand dunes can be seen under a light blue sky. Sunlight shines from the upper left, casting long shadows. The entire environment is dry and open, with no other vegetation, while tiny grains of sand can be seen floating in the air, creating a typical desert landscape.''

Output:

[

    {"question": "What kind of natural environment and overall landscape does this image show?", "answer": "The image shows a dry and open desert environment featuring a green cactus about 60 centimeters tall with three pink flowers on top, surrounded by gray rocks, gently rolling sand dunes in the distance, a light blue sky, and sunlight shining from the upper left, forming a typical desert landscape."},
    
    {"question": "What are the height and shape of the cactus in the image?", "answer": "Approximately 60 centimeters tall, with a green cylindrical stem"},
    
    {"question": "What features are present at the top of the cactus?", "answer": "Three pink flowers with outward-spreading petals and yellow stamens"},
    
    {"question": "What features are present on the surface of the cactus?", "answer": "Prominent spines"},
    
    {"question": "What is the color and surface characteristic of the rock near the cactus?", "answer": "Gray, with a rough surface and cracks"},
    
    {"question": "What is the terrain in the distance?", "answer": "Gently rolling sand dunes"},
    
    {"question": "What color is the sand in the desert environment?", "answer": "Light yellow"},
    
    {"question": "From which direction does the sunlight shine?", "answer": "From the upper left"},
    
    {"question": "What visible phenomenon is present in the air?", "answer": "Tiny grains of sand floating in the air"}
    
]
\end{mdframed}

\subsubsection{Prompt for Hierarchical Semantic Analysis}

The following prompt is used to extract hierarchical semantic information from images for semantic grouping and data construction.

\begin{mdframed}
\small
\ttfamily
Please conduct an in-depth analysis of the input image, extract hierarchical semantic information from its ``static'' content, and ultimately output a strict JSON object.

[Hierarchical Semantic Structure Analysis (Content Understanding)]

1. Core Objective:  

Extract key information at the conceptual, categorical, and functional levels from the image, to support the construction of a semantically related image database.

2. General Rules:

(1) Understand first, then abstract:  

Fully understand the image content (including any text), then summarize and abstract concrete content into concepts, categories, and functions.

(2) Concept first:  

Always focus on:  

-- what it is (theme/object)  

-- what it is about (concept/knowledge)  

-- why it exists (function/purpose)

(3) Avoid surface features:  

Colors, angles, backgrounds, specific fonts, decorative layouts, and non-essential visual styles should not be treated as core content in the first six fields. These belong to low-value surface information.

3. Fields and Instructions:

(1) Type Theme (required):  

The macro-level presentation form.  

Examples: Nature and Environment; People and Society; Virtual and Art; Academic Problem; Knowledge Introduction/Document; Screenshot/Interface; Others.

(2) Domain Direction (required):  

The broader discipline, industry, or application domain.  

Examples: Natural Science/Geography; Business and Economics; Education and Training/Mathematics; History and Culture; Information Technology.

(3) Semantic Subcategory (required):  

A specific topic, problem type, or content theme under the Domain Direction.  

Examples: Coastal Landforms; Technology Company Introduction; Quadratic Function Evaluation; Renaissance Painting.

(4) Core Objects:  

The main entities or key information items that act as information carriers in the image.  
Use nouns or noun phrases.

(5) Core Concepts:  

Abstract knowledge, principles, or ideas conveyed through the core objects.

(6) Function or Role:  

The primary static purpose or role of the image.

(7) Distinguishing Key Information:  

Content features used to differentiate images within the same Semantic Subcategory.

(8) Low-Value Surface Information:  

Visual or surface-level features that should not be used as matching criteria.

[Overall Output JSON Format]
\{

    "hierarchical\_semantic\_information": \{

    "type\_theme": "",
    
    "domain\_direction": "",
  
    "semantic\_subcategory": "",
    
    "core\_objects": [],
    
    "core\_concepts": [],
    
    "function\_or\_role": "",
    
    "distinguishing\_key\_information": [],
    
    "low\_value\_surface\_information": []

  \}
\}

Now, please analyze the given image and output the complete JSON result in English.  
Strictly follow the format above and do not include any additional explanations.

\end{mdframed}

\subsubsection{Prompt for Image-Pair Caption Generation}

The following prompt is used to generate joint captions for semantically related image pairs.

\begin{mdframed}
\small
\ttfamily
You will be given two images. Please generate a single, complete, logically coherent English caption that provides a joint description of both images.

[Core Principles and Scope of Information]

Your task is to conduct a high-level comparative and relational analysis of the two images while ensuring extremely high information density and completeness.

1. Focus on the core; ignore non-essential elements:  

The description must focus entirely on the primary content that conveys core information, such as main text, titles, key data, central arguments, or critical facts.  

For webpage or interface screenshots, strictly ignore and never mention:

-- navigation bars, menus, footer areas  

-- registration records, copyright notices, auxiliary modules  

-- statistical or interactive elements (e.g., view counts, edit counts, feedback links)  

-- layout features such as ``Table of Contents,'' ``Gallery,'' or ``References''  

It is strictly forbidden to treat superficial structural similarities (e.g., similar page layout) as meaningful content similarities. All comparisons must be based on substantive content.

2. Detailed informational grounding:  

-- Provide detailed transcription of key textual information; avoid vague expressions such as ``for example'' or ``etc.''  

-- Clearly describe logical relationships among core elements, without referencing container-like section names  

3. Objective description:  

Only describe observable facts. Do not speculate or evaluate.

[Caption Generation Structure Requirements]

1. High-level overview:  

Identify the macro-level theme, field, or concept shared by both images.

2. Description of Image A:  

Introduce naturally (e.g., ``One of the materials involves...'') and provide a complete and detailed description of:

-- core subject or theme  

-- key textual content  

-- main aspects of the content  

-- (if applicable) its role or significance  

3. Description of Image B:  

Introduce similarly (e.g., ``The other material involves...'') and provide an equally detailed description.

4. Comparative and relational analysis:  

-- Core similarities: identify shared subject matter or conceptual domain  

-- Key differences: analyze differences in subject, context, roles, themes, or audience, grounded in content  

-- Complementarity: explain how differences together form a richer structure or understanding  

5. Conclusion:  

Briefly state the significance of the comparison for understanding the broader theme.

[Language and Output]

-- Use objective, concise, and information-dense English  

-- Do not refer to ``first image'' or ``second image''  

-- Do not mention layout, colors, or formatting  

-- Output a single coherent paragraph only  

Note: Output only the generated caption. Do not add any explanation.

\end{mdframed}

\subsubsection{Prompt for Text-Based Knowledge Extraction}

The following prompt is used to extract structured semantic information from text descriptions (e.g., image captions or multi-turn QA).

\begin{mdframed}
\small
\ttfamily
You are a strict text-based knowledge extraction expert.

Task:
You will be given a piece of text (either an image description or multi-turn question--answer content). You must extract information strictly from this text only.  

Do not add any information that is not explicitly stated in the text.  

Do not infer from the image.  

Do not use any background knowledge.

You need to extract two types of knowledge: Fact and Abstract.

Definitions:

Fact:

-- Objectively verifiable information explicitly described in the text  

-- Includes: existence of objects, attributes (such as color, size, position), actions, quantities, and relationships  

-- Characteristics: concrete, objective, directly observable  

Examples:  

``The cat is black''  

``The cat is sitting on the windowsill''  

``There is sunlight outside the window''  

Abstract:

-- Subjective feelings, emotions, atmosphere, evaluations, or background information explicitly expressed in the text  

-- Includes: emotional states (e.g., relaxed, happy), atmosphere (e.g., warm, peaceful), evaluations (e.g., beautiful, comfortable), speculation (e.g., ``might be \ldots''), purpose (e.g., ``suitable for \ldots'')  

-- Characteristics: abstract, subjective, requires contextual understanding  

Examples:  

``The cat looks relaxed''  

``It gives a warm and cozy feeling''  

``It might be a residence''  

Rules:

1. Fact:

-- Must be explicitly stated in the text; no inference is allowed  

-- Do not include any attributes, objects, or scenes that are not mentioned in the text  

-- Each Fact should be a JSON object with:

   "fact": atomic fact (short description)  
   
   "level": "L1"  

2. Abstract:

-- Must be explicitly expressed in the text  

-- Do not add any emotions, atmosphere, or interpretations that are not present in the text  

-- Do not generate inferred abstract information  

-- Output as a list of short sentences  

Important Constraints:

-- If an attribute (e.g., color, shape, position) is not mentioned, do not add it  

-- Do not perform visual inference  

-- Do not add common-sense knowledge  

-- Only extract information explicitly present in the text  

Output Format:

-- Output JSON only; do not include any explanation  

-- The JSON must contain two keys: "Fact" and "Abstract"  

-- If either is empty, output an empty list  

\end{mdframed}

\subsection{Qualitative Examples: Paired Image Captions}

\begin{mdframed}
\small

\textbf{Input Images}

\begin{center}
\includegraphics[width=0.45\linewidth]{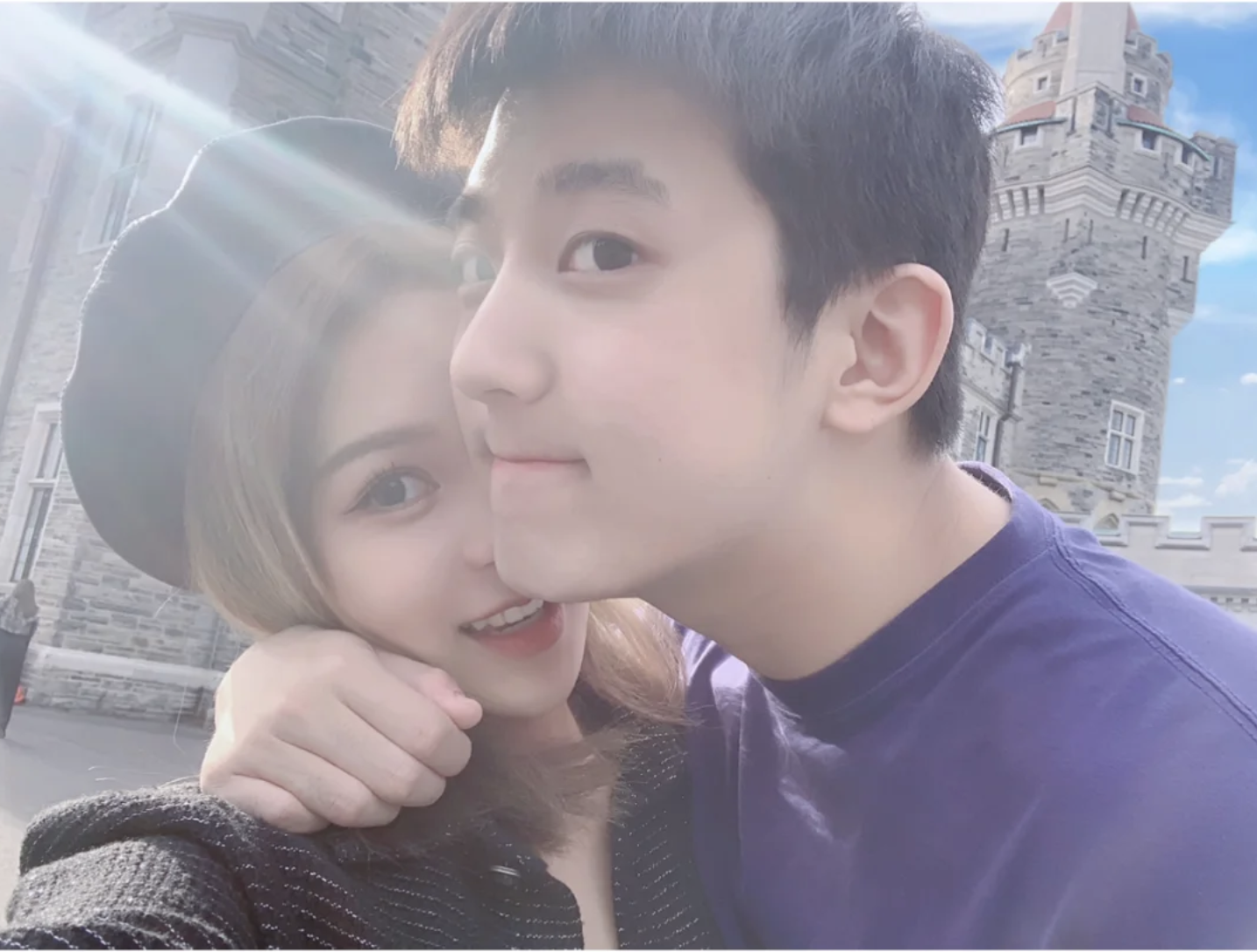}
\hfill
\includegraphics[width=0.45\linewidth]{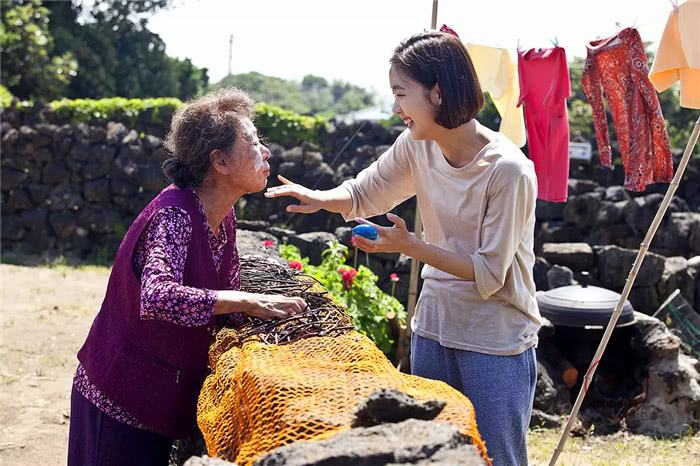}

\end{center}

\vspace{4pt}

\textbf{Single-Image Captions}

\textit{Image A:} \\

This image shows two young people outdoors, who appear to be smiling and leaning close to each other while taking a selfie. In the background, there is a building that resembles a castle or tower, suggesting that they may be at a tourist attraction or historical site. Sunlight enters from the right side of the image, creating a bright halo effect and adding a warm atmosphere to the scene.

\vspace{3pt}

\textit{Image B:} \\

‘Goblin Bride’ Kim Go-eun teams up with Korean film queen—‘Canola’ overflows with tear-jerking grandmother-granddaughter affection.

\vspace{6pt}

\textbf{Generated Pair Caption}

\ttfamily

These two images both focus on interpersonal interaction and emotional communication, illustrating the diversity and depth of human relationships across different contexts.

One of the images depicts an intimate interaction between a young man and a young woman outdoors. In the scene, the two stand in front of a castle-like structure with a historical atmosphere, set against a background of blue sky and sunlight, creating a warm and romantic ambiance. The man is dressed in dark clothing, while the woman wears a black hat. They stand close to each other, with the woman resting her arm on the man’s shoulder. Their expressions appear natural and affectionate, suggesting they are enjoying each other’s company. This scene highlights the emotional connection and relaxed, pleasant atmosphere characteristic of intimate relationships.

The other image portrays an interaction between a young woman and an older woman in an outdoor setting. The background suggests a rural environment, with clothes hanging on a line, surrounded by stone walls and plants, giving a simple and natural impression. The young woman, dressed in a light-colored top, holds a small object and appears to be communicating with the older woman, who is focused on her work. There is a sense of warmth and familiarity in their interaction. This scene emphasizes intergenerational communication and care, reflecting the emotional bond between individuals of different age groups.

\#\#\# In-depth Comparative and Relational Analysis:

- **Core similarities**: Both images center on human interaction and the exchange of emotions, presenting different types of relationships. Whether in the form of intimacy characterized by romance and emotional closeness, or intergenerational interaction characterized by care and communication, both convey warmth and connection in human relationships.

- **Key differences and contrasts**:

  - **Type of relationship**: The first image depicts an intimate relationship, emphasizing romance and emotional closeness, while the second image represents an intergenerational relationship, emphasizing care and depth of communication.

  - **Contextual setting**: The first image features a castle-like background with historical connotations, evoking a sense of romance and idealization. In contrast, the second image is set in a rural environment, which appears simple and natural, highlighting the authenticity of everyday emotional interactions.

  - **Emotional expression**: In the first image, emotions are conveyed through body language and facial expressions, expressing intimacy and joy. In the second image, emotions are expressed more subtly through actions and posture, reflecting care and respect.

- **Internal connections and complementarity**: 

Together, the two images reveal the diversity and richness of human relationships. The direct emotional expression found in intimate relationships contrasts with the depth and continuity of emotion found in intergenerational relationships. This comparison not only highlights the distinctive emotional characteristics of different relationship types, but also reflects the universality and diversity of human emotions.

\#\#\# Conclusion:

Through different contexts and relationship types, these two images jointly construct a rich portrayal of human relationships, revealing the diversity and depth of emotional communication across different settings. Whether through romantic intimacy or intergenerational care, both illustrate the warmth and connection inherent in human emotions, offering a multidimensional perspective on the complexity and richness of interpersonal relationships.

\normalfont

\end{mdframed}

\subsection{Qualitative Examples: Multi-Image Interleaved Captions}

\begin{mdframed}
\small

\textbf{Input Images}

\begin{center}

\includegraphics[width=0.3\linewidth]{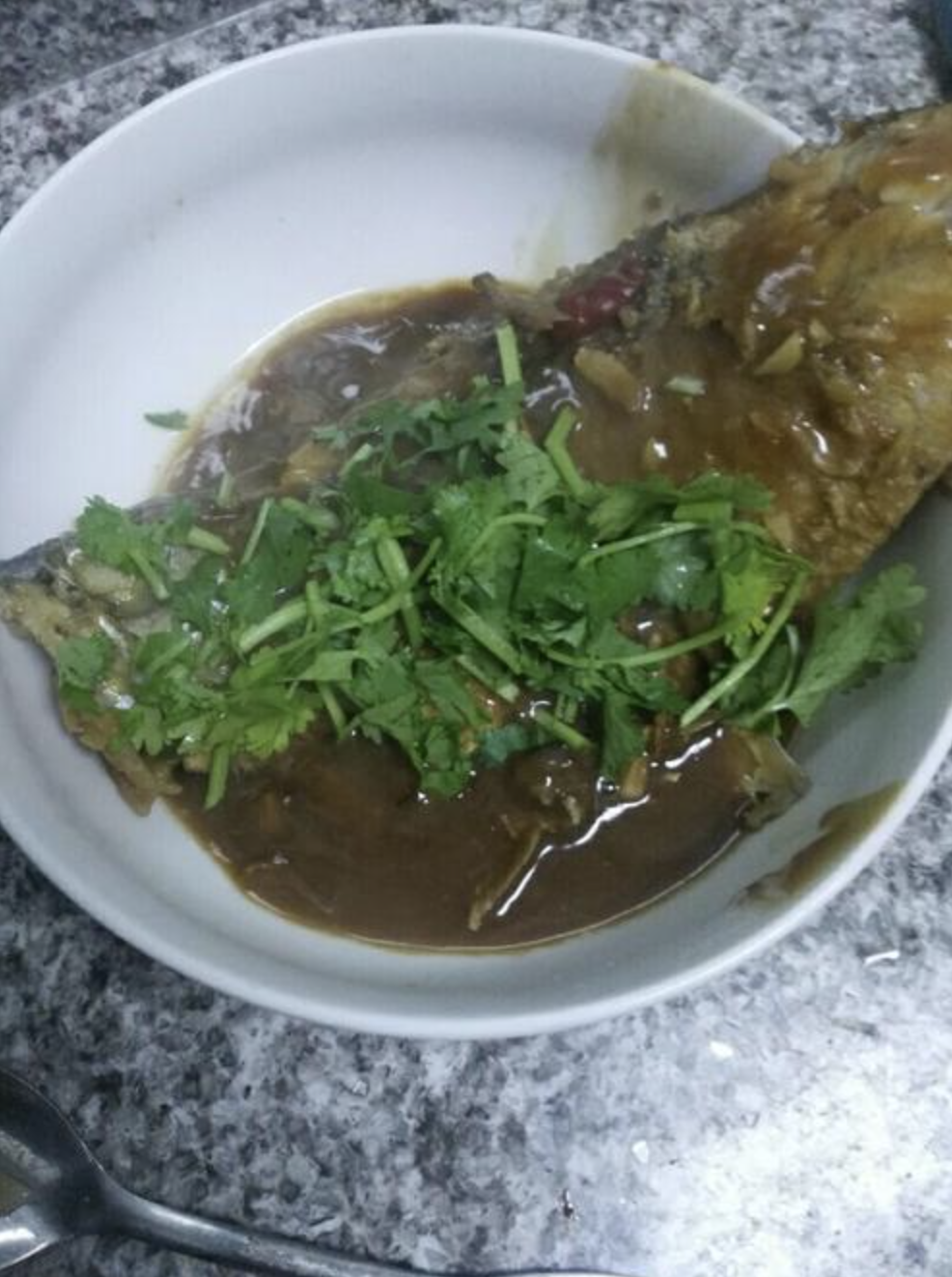}

\vspace{6pt}

\includegraphics[width=0.45\linewidth]{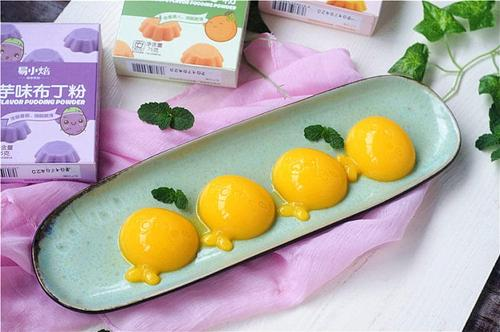}
\hfill
\includegraphics[width=0.5\linewidth]{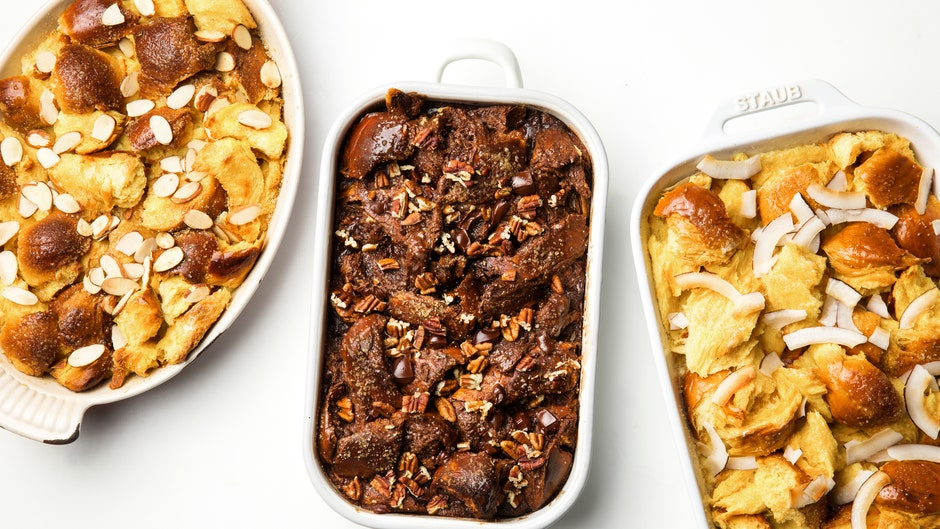}

\vspace{6pt}

\includegraphics[width=0.45\linewidth]{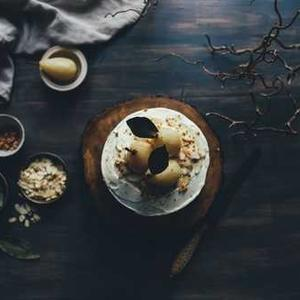}
\hfill
\includegraphics[width=0.3\linewidth]{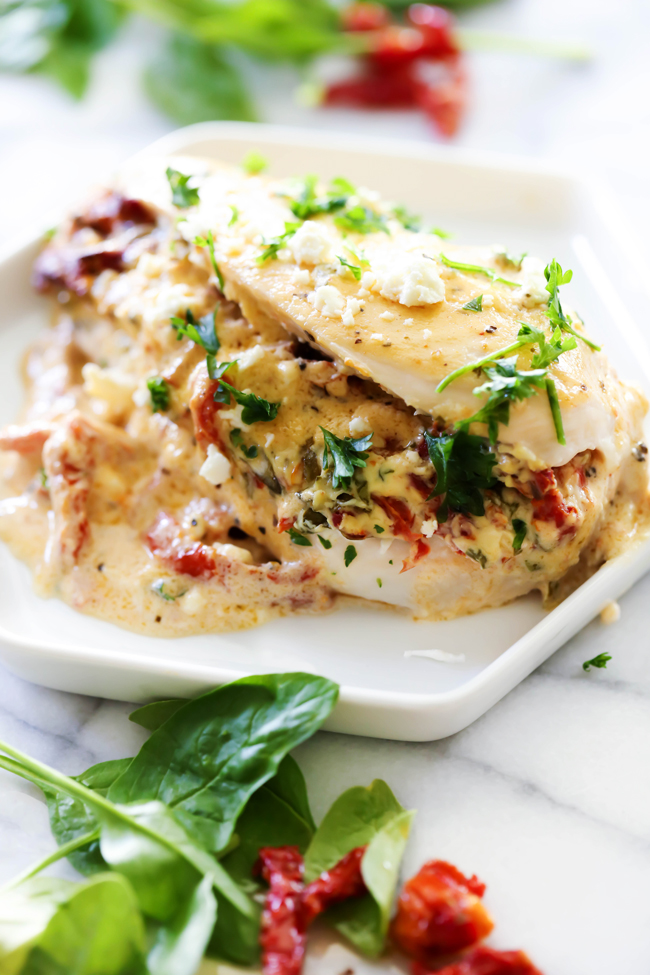}
\end{center}

\vspace{6pt}

\textbf{Generated Multi-Image Descriptions}

\ttfamily
In a warm and cozy family kitchen, a mother is preparing a delicious dinner.

First, she serves the opening dish: a braised fish. The fish is golden in color, with a crispy exterior and rich, flavorful sauce, garnished with fresh cilantro and emitting an inviting aroma. This dish symbolizes the warmth and togetherness of family, bringing a sense of home whether on a busy evening or a leisurely afternoon. <Image\_1>

Next, the mother presents a delicate dessert—taro pudding. The puddings are neatly arranged on a long plate, brightly colored and uniformly golden, with smooth surfaces and a sweet fragrance. Decorative green leaves and a soft pink cloth beside the plate add a touch of freshness and elegance. <Image\_2> Seeing these delightful puddings, the children’s eyes light up as they eagerly anticipate tasting them.

The third dish is a variety of bread puddings, carefully prepared as the main course. Three different flavors are placed in separate baking trays: a golden pudding topped with almond slices, a deep brown pudding sprinkled with crushed walnuts, and another golden pudding decorated with white coconut flakes. Each variation releases its own distinct aroma, allowing every family member to find a favorite. <Image\_3> The surfaces are baked to a perfect golden finish, making them especially appetizing.

Finally, to complete the dinner, the mother prepares a beautifully decorated cake for dessert. The cake is covered with smooth white frosting, adorned with dark chocolate pieces, and accompanied by nuts and small plates that add visual depth. The dark tablecloth in the background contrasts with the white cake, creating an elegant appearance. <Image\_4> The dining table is filled with a variety of delicious foods, each bite bringing a sense of happiness.

To conclude the meal, the mother serves an impressive dish of lasagna. This hearty main course is cut into neat portions and placed on a white plate. Each slice reveals distinct layers of filling, including cheese, vegetables, white meat, and rich creamy sauce, topped with herbs and a light sprinkling of cheese for added flavor. Fresh green lettuce and roasted tomatoes are arranged beside it, enriching the overall color palette. <Image\_5> The family gathers around the table, enjoying the carefully prepared dinner, with satisfied smiles on their faces.

This dinner not only satisfies the family’s taste buds but also conveys the mother’s love and care. Through these five dishes, a warm family moment unfolds, with every detail reflecting the meaning of home.
\normalfont

\end{mdframed}

%%%%%%%%%%%%%%%%%%%%%%%%%%%%%%%%%%%%%%%%%%%%%%%%%%%%%%%%%%%%%%%%%%%%%%%%%%%%%%%
%%%%%%%%%%%%%%%%%%%%%%%%%%%%%%%%%%%%%%%%%%%%%%%%%%%%%%%%%%%%%%%%%%%%%%%%%%%%%%%

\end{document}